\documentclass{article} 
\usepackage{iclr2027_conference,times}

\usepackage{amsmath,amsfonts,bm}

\def\eqref#1{equation~\ref{#1}}

\def\1{\bm{1}}

\DeclareMathAlphabet{\mathsfit}{\encodingdefault}{\sfdefault}{m}{sl}
\SetMathAlphabet{\mathsfit}{bold}{\encodingdefault}{\sfdefault}{bx}{n}

\usepackage{booktabs}
\usepackage{graphicx}
\usepackage{hyperref}
\usepackage{url}
\usepackage{multirow}
\usepackage[table]{xcolor}
\title{What Generation Knows: Unlocking Fine-Grained Understanding in Unified Multimodal Models}

\author{%
\makebox[\textwidth][c]{%
{\normalfont\mdseries
\begin{tabular}{@{}c@{}}
Tengju Ru$^{1,4,*}$ \quad Weitong Lian$^{1,*}$ \quad Zecong Tang$^{2,*}$ \quad Zhiheng Liu$^{1}$ \quad Lingyi Meng$^{1}$ \\[3pt]
Haoran Li$^{2}$ \quad Kaixuan Wang$^{2}$ \quad Bohao Li$^{3}$ \quad Xihui Liu$^{2,4,\dagger}$ \\[5pt]
{\footnotesize $^{1}$ Zhejiang University \quad $^{2}$ The University of Hong Kong} \\[2pt]
{\footnotesize $^{3}$ The Chinese University of Hong Kong, Shenzhen \quad $^{4}$ Shenzhen Loop Area Institute} \\[4pt]
{\footnotesize $^{*}$ Equal contribution. \quad $^{\dagger}$ Corresponding author.} \\[3pt]
{\footnotesize
\href{https://github.com/LRuca/GAFU}{\texttt{https://github.com/LRuca/GAFU}}}
\end{tabular}%
}%
}%
}

\iclrfinalcopy 
\begin{document}

\maketitle

\fancyhead{}
\renewcommand{\headrulewidth}{0pt}

\begin{abstract}
Fine-grained visual understanding requires distinguishing subtle differences in local attributes and spatial relations, yet remains challenging for existing vision-language models.
Unified multimodal models (UMMs) offer a new opportunity through generation pathways that preserve rich visual information complementary to understanding.
However, existing generation-assisted understanding methods mainly rely on generative supervision or explicit generation and have not systematically explored direct use of native generation-side representations.
To address this gap, we systematically analyze what generation-side representations offer for fine-grained understanding, finding that they are more sensitive to fine-grained semantic changes, capture task-relevant changes missed by understanding-side representations, and depend on correct spatial organization for this advantage.
Guided by these findings, we design GAFU, a post-training framework that unlocks native generation-side information for fine-grained understanding. 
It compresses generation-side visual representations into compact \(G\) tokens that preserve fine-grained visual details, enabling task-conditioned retrieval by the understanding pathway.
Experiments show that GAFU consistently outperforms SFT baselines trained on the same data across diverse visual understanding tasks and UMM architectures, with maximum relative improvements of 20.00\%, 17.94\%, and 8.62\% on VStar-Bench, OmniSpatial, and VSR-Bench, respectively.
Overall, GAFU converts complementary information from the generation pathway into gains in fine-grained visual understanding, providing an effective path toward synergy between generation and understanding in UMMs.

\end{abstract}

\section{Introduction}

In recent years, vision-language models (VLMs) have made substantial progress in visual understanding \citep{li2024llavaonevisioneasyvisualtask,bai2025qwen3vltechnicalreport,zhu2025internvl3exploringadvancedtraining}.
However, this progress remains uneven between high-level semantic understanding and fine-grained visual understanding: while VLMs are comparatively strong in high-level semantic understanding, they remain substantially less reliable on fine-grained tasks involving objects, attributes, counting, and spatial relations \citep{tong2024eyeswideshutexploring,fu2024blinkmultimodallargelanguage,kamoi2025visonlyqalargevisionlanguage}.
Existing methods mainly improve fine-grained visual understanding through higher-fidelity visual encoding and more information-preserving vision--language interfaces \citep{you2023ferretrefergroundgranularity,li2024monkeyimageresolutiontext,tong2024cambrian1fullyopenvisioncentric,zhang2025llavauhdv2mllmintegrating}, but still rely on language-aligned vision encoders that compress visual content into limited representations before language modeling, so that fine-grained evidence lost upstream becomes unavailable downstream.

UMMs jointly model multimodal understanding and generation within a single model \citep{chameleonteam2025chameleonmixedmodalearlyfusionfoundation,zhao2026unifiedmultimodalunderstandinggeneration}.
Compared with VLMs that rely primarily on language-aligned visual representations, the generation pathway in UMMs typically preserves richer structural information, spatial relations, and local details \citep{Wu_2025_CVPR,Qu_2025_CVPR,ma2025neurips-unitok}.
Meanwhile, the understanding and generation pathways share knowledge and mutually adapt during joint modeling, providing a basis for complementary use of the two types of visual information \citep{deng2025emergingpropertiesunifiedmultimodal,kang2026transferabilityunderstandinggenerationunified}.
However, fine-grained visual understanding remains challenging for UMMs, and it remains unclear whether and under what conditions generation-side representations provide additional evidence for fine-grained understanding, as well as how such evidence can be effectively exploited by the understanding pathway \citep{liu2026unisonbenchmarkingunifiedmultimodal,wang2026quantifyinggapunderstandinggeneration,wen2026unig2ubenchunifiedmodelsadvance}.

Generation-assisted understanding in UMMs mainly follows two directions.
One line uses generative supervision to shape visual representations or internal hidden states to encode richer fine-grained visual information \citep{wang2024diffusionfeedbackhelpsclip,luo2025deemdiffusionmodelsserve,wang2024reconstructivevisualinstructiontuning}.
However, these methods do not directly examine or exploit information in native generation-side representations.
Their heuristic supervision targets limit task-conditioned visual evidence selection, while differing representational requirements of understanding and generation may introduce representational conflicts \citep{ma2025genhancerimperfectgenerativemodels,chen2025uniheterogenerationenhanceunderstanding,su2026generationenhancesunderstandingunified}.
Another line explicitly generates visual intermediate states at inference time and uses them for downstream understanding \citep{su2025thinkingimagesmultimodalreasoning,li2025imaginereasoningspacemultimodal}.
However, this approach externalizes generation-side information into limited intermediate states, with some methods re-encoding the outputs into the understanding space and reintroducing the visual encoding bottleneck of the understanding pathway \citep{cheng2025visualthoughtsunifiedperspective,yang2026imaginevisualthinkingunified,hao2026clearunlockinggenerativepotential,cheng-etal-2026-omni}.
Explicit generation introduces inference overhead and may propagate generation errors into downstream understanding \citep{meng2026imugbenchbenchmarkingunifiedmultimodal}.

To address these limitations, we first examine how generation-side representations contribute to fine-grained understanding.
We focus on whether they contain visual cues beyond those captured by the understanding pathway and on the factors that determine this advantage.
We find that generation-side representations respond more strongly to subtle semantic changes, capture changes missed by understanding-side representations, and rely on  spatial organization to retain this advantage.

Guided by these findings, we propose GAFU, a Generation-Assisted Fine-grained Understanding post-training framework that directly exploits native generation-side representations.
To preserve fine-grained information, GAFU transforms native generation-side representations into compact \(G\) tokens through compression, spatial binding, and relation alignment.
We then use the evolving text states to selectively retrieve task-relevant evidence from \(G\) and feed it back into the understanding pathway.
We further use pairwise preference learning to encourage reliance on sample-relevant generation-side evidence while preserving the original understanding capability.
We evaluate GAFU across UMMs with diverse architectures and scales on general, spatial, and domain-specific visual understanding tasks.
GAFU outperforms same-data SFT in all 15 model--benchmark combinations, with substantial average gains of 3.56--5.98 percentage points across the five benchmarks.

Our contributions can be summarized as follows:

\begin{enumerate}
\item We systematically study understanding- and generation-side visual representations in UMMs, revealing generation-side advantages in fine-grained semantic sensitivity, representational complementarity, and spatial dependence.

\item Based on these findings, we propose GAFU, which enables the understanding pathway to exploit complementary generation-side information through representation compression and alignment, task-conditioned injection, and pairwise preference learning.

\item We validate GAFU across diverse UMM architectures and visual understanding tasks, outperforming same-data SFT in all 15 settings by 3.56--5.98 percentage points on average.
\end{enumerate}

\section{Related Work}

\subsection{Unified Multimodal Models}
UMMs jointly model multimodal understanding and visual generation within a single model through autoregressive, diffusion-based, or hybrid modeling paradigms \citep{wang2024emu3nexttokenpredictionneed,xie2025showosingletransformerunify,you2026lladaoeffectivelengthadaptiveomni,zhang2026pelicanunify10unifiedembodied}.
Early UMMs often relied on a unified dense backbone with shared visual representations \citep{chameleonteam2025chameleonmixedmodalearlyfusionfoundation,wang2024emu3nexttokenpredictionneed}, while later designs increasingly specialize the visual pathways for understanding and generation within a unified framework \citep{Wu_2025_CVPR,fu2026lanceunifiedmultimodalmodeling}.
This evolution reflects different representation needs: understanding relies more on high-level semantics, whereas generation requires richer visual details, motivating many UMMs to use language-aligned encoders such as CLIP or SigLIP for understanding and VAE- or VQ-VAE-based encoders for generation \citep{Wu_2025_CVPR,wu2025harmonizingvisualrepresentationsunified,fu2026lanceunifiedmultimodalmodeling}.
Despite this progress, architectural unification has not yet translated into effective synergy between understanding and generation \citep{shi2026realunifyunifiedmodelstruly}, with persistent capability gaps \citep{wang2026quantifyinggapunderstandinggeneration} and limited generation-to-understanding transfer \citep{wen2026unig2ubenchunifiedmodelsadvance,zhu2026doesvisualgenerationhelp}.

\subsection{Understanding--Generation Collaboration in Unified Multimodal Models}
Collaboration between understanding and generation in UMMs mainly follows two directions: understanding-guided generation and generation-assisted understanding.
Understanding-guided generation uses understanding-side information for generation conditioning \citep{dong2024dreamllmsynergisticmultimodalcomprehension,chen2025blip3ofamilyfullyopen,xie2026reconstructionalignmentimprovesunified}, task planning \citep{zeng2026drawinmindrebalancingdesignerpainterroles,liu2026breakingdualbottlenecksevolving,li2026cococodecottexttoimage}, and output feedback \citep{jin2026srumfinegrainedselfrewardingunified,huang2026alphagrpounlockingselfreflectivemultimodal,liu2026breakingdualbottlenecksevolving}, forming a relatively mature paradigm.
In contrast, generation-assisted understanding has received comparatively less systematic attention.
It mainly relies on generative supervision to shape visual representations or internal states \citep{wang2024diffusionfeedbackhelpsclip,ma2025genhancerimperfectgenerativemodels,wang2024reconstructivevisualinstructiontuning}, which can constrain representation learning through indirect objectives, or on generated visual intermediate states \citep{li2025imaginereasoningspacemultimodal,tong2026reversingflowgenerationtounderstandingsynergy,hao2026clearunlockinggenerativepotential}, which introduce a costly and lossy intermediate generation process \citep{meng2026imugbenchbenchmarkingunifiedmultimodal}.
However, the role of native generation-side representations in understanding remains underexplored, particularly whether they provide complementary evidence, what enables such advantages, and how they can be used effectively.

\section{Methodology}

We first examine whether generation-side representations provide additional evidence for fine-grained understanding and what underlies any such advantage, and then develop a post-training framework to make this information effectively usable by the understanding pathway.

\subsection{Systematic Analysis of Visual Representations}
\label{subsec:systematic-analysis}

We compare generation- and understanding-side visual representations along three dimensions: fine-grained semantic sensitivity, representational complementarity, and spatial dependence, to characterize the generation-side advantage and its underlying structure.
We study six UMMs---BAGEL~\citep{deng2025emergingpropertiesunifiedmultimodal},
Janus-Pro~\citep{chen2025janusprounifiedmultimodalunderstanding},
Lance~\citep{fu2026lanceunifiedmultimodalmodeling},
Show-o2-7B~\citep{xie2025showo2improvednativeunified},
BLIP3-o-4B~\citep{chen2025blip3ofamilyfullyopen},
and DreamLLM-7B~\citep{dong2024dreamllmsynergisticmultimodalcomprehension}---and compare each model's generation-side representation with its corresponding understanding-side visual representation.
Our analysis uses 2,084 minimally changed VisMin image pairs spanning \texttt{object}, \texttt{attribute}, \texttt{count}, and \texttt{relation}.
Figure~\ref{fig:systematic-study} summarizes the three analyses.

\subsubsection{Fine-Grained Semantic Sensitivity}

For generation-side representations to complement fine-grained understanding, they should first be sensitive to subtle semantic changes critical to these tasks.
Let \(R_G\) and \(R_U\) denote generation-side and understanding-side representations, respectively, and let \(\bar{R}(I,p)\) denote the feature at position \(p\) after aligning representation \(R\in\{R_G,R_U\}\) of image \(I\) to a \(16\times16\) grid.
Since representation responses may also reflect low-level appearance variation, for each edit $I\rightarrow I_v$ we construct a control $I_n$ within the same edited region with matched linear-RGB RMS change magnitude.
Let \(\mathcal{W}_{2\times2}\) denote all contiguous \(2\times2\) windows on the aligned grid. We define the position-wise response \(d_R\), raw local response \(\rho_R\), and log-transformed sensitivity \(S_R\) as
\begin{equation}
\begin{aligned}
d_R(I_a,I_b;p)
&=\max\!\left(0,\,1-\cos\!\left(\bar{R}(I_a,p),\bar{R}(I_b,p)\right)\right),\\
\rho_R(I_a,I_b)
&=\textstyle\max_{\Omega\in\mathcal{W}_{2\times2}}\tfrac14\sum_{p\in\Omega}d_R(I_a,I_b;p),\\
S_R(I_a,I_b)
&=\log\!\left(\rho_R(I_a,I_b)+10^{-10}\right).
\end{aligned}
\label{eq:representation-sensitivity}
\end{equation}
For sample \(i\), we define the control-corrected sensitivity and its domain-level mean as
\begin{equation}
C_R^i=S_R(I_i,I_{v,i})-S_R(I_i,I_{n,i}),
\qquad
\bar{C}_R^{(d)}=\tfrac{1}{N_d}\textstyle\sum_{i\in\mathcal{D}_d}C_R^i.
\label{eq:control-corrected-sensitivity}
\end{equation}
where \(\mathcal{D}_d\) denotes the samples in domain \(d\) and \(N_d=|\mathcal{D}_d|\). A larger \(\bar{C}_R^{(d)}\) indicates stronger semantic sensitivity after controlling for matched low-level variation.
Details of the control construction, absolute sensitivity values, and resolution analysis are provided in Appendix~\ref{app:representation-sensitivity}.

As shown in Figure~\ref{fig:systematic-study}(a), generation-side representations achieve higher control-corrected sensitivity than understanding-side representations in 24 model--domain combinations, establishing an advantage in fine-grained semantic sensitivity.
On average across all six models, this advantage is smallest for \texttt{count} and larger for \texttt{attribute}, \texttt{object}, and \texttt{relation}, indicating that generation-side representations provide stronger advantages for changes that depend on local details and spatial structure than for those relying on cross-instance aggregation.
The generation-side advantage also remains positive across all tested input resolutions, confirming its robustness to input scale.

\subsubsection{Complementarity Analysis}
Having established a sensitivity advantage, we next ask whether this extra responsiveness captures changes missed by understanding-side representations, since both may still respond strongly to the same visual changes.
For each sample \(i\), using the raw local response \(\rho_R\) defined above, we define the strong-response indicator as
\begin{equation}
q_R^i
=
\mathbb{I}\!\left[
\rho_R(I_i,I_{v,i})
>
2\,\rho_R(I_i,I_{n,i})
\right],
\qquad
R\in\{R_G,R_U\}.
\label{eq:strong-response}
\end{equation}
For each domain \(d\), we report the fraction where only the generation-side representation is strongly responsive as \(P_{G\text{-only}}^{(d)}\), and the fraction where only the understanding-side is strongly responsive as \(P_{U\text{-only}}^{(d)}\). Their difference, \(\Delta_{\mathrm{comp}}^{(d)}=P_{G\text{-only}}^{(d)}-P_{U\text{-only}}^{(d)}\), measures the net generation-side advantage.

\begin{figure*}[t]
\centering
\includegraphics[width=\textwidth]{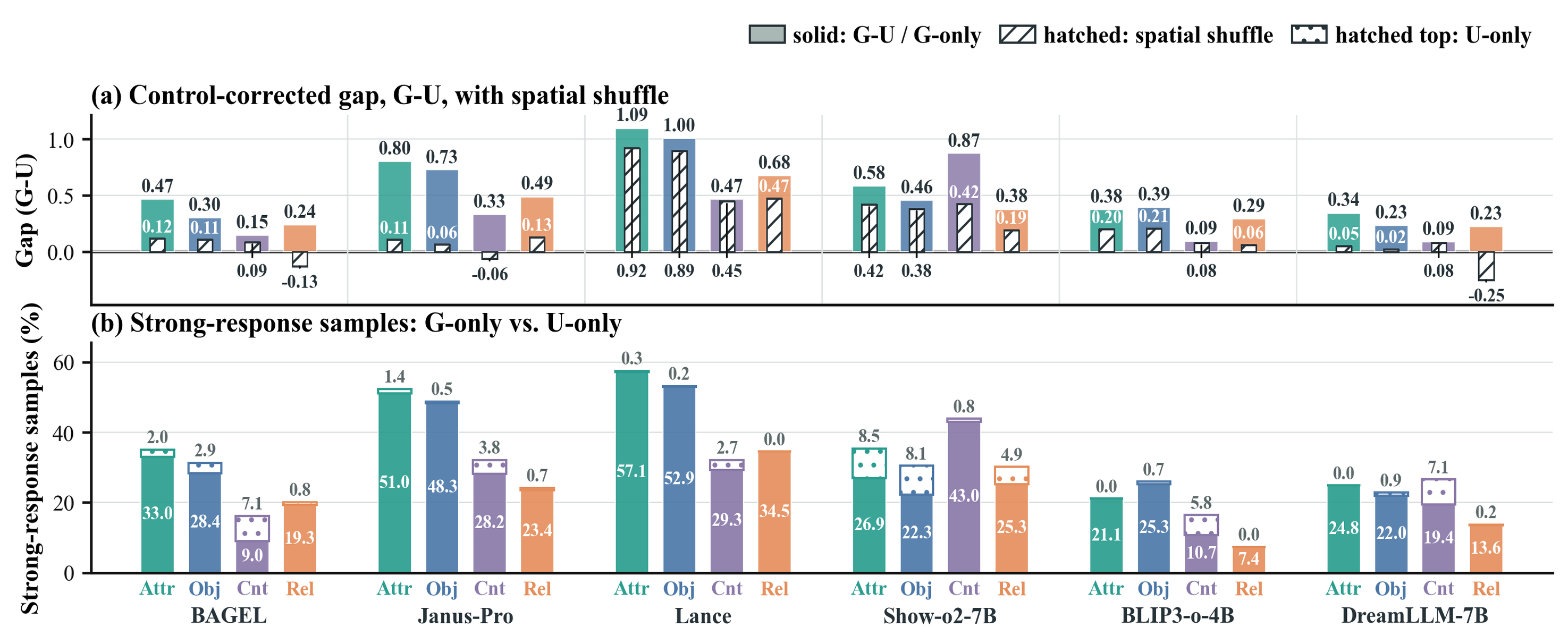}
\caption{Control-corrected generation--understanding sensitivity gaps and strong-response complementarity across six models and four domains. (a) Solid bars show the unpermuted $G-U$ gap, while hatched bars show the gap after spatial shuffle. (b) Stacked bars show the fractions of samples with generation-side-only and understanding-side-only strong responses.}
\label{fig:systematic-study}
\end{figure*}

As shown in Figure~\ref{fig:systematic-study}(b), \(P_{G\text{-only}}^{(d)}\) exceeds \(P_{U\text{-only}}^{(d)}\) in all 24 model--domain combinations, establishing systematic representational complementarity across all six models and four domains.
This consistent asymmetry shows that the generation-side advantage extends beyond higher overall sensitivity: generation-side representations also capture task-relevant visual changes missed by understanding-side representations.
Detailed statistics are provided in Appendix~\ref{app:representation-sensitivity}.

\subsubsection{Spatial Dependence}

Despite this complementarity, the conditions supporting this advantage remain unclear.
Since generation-side representations preserve rich spatial structure, we further test whether this advantage depends on spatial organization.
We apply a fixed permutation over a \(64\times64\) grid of image blocks before encoding and compare it with the unpermuted condition. We recompute control-corrected sensitivities under permutation. The detailed values are provided in Appendix~\ref{app:representation-sensitivity}.

As shown in Figure~\ref{fig:systematic-study}(a), spatial permutation reduces the generation--understanding sensitivity gap in all 24 model--domain combinations, demonstrating that spatial organization is a systematic source of the generation-side advantage rather than a model- or task-specific effect.
The role of spatial organization is most direct for \texttt{relation}, where spatial layout defines the semantic change. Importantly, the substantial reductions for \texttt{attribute} and \texttt{object} show that this dependence is not limited to explicitly relational changes, but extends to fine-grained evidence whose usefulness depends on preserving its binding to object locations and surrounding structure.
Notably, \texttt{count} exhibits the smallest gap reduction on average across models, revealing a clear task-dependent pattern: the generation-side advantage is strongest when fine-grained evidence must remain locally grounded and spatially organized, and weaker when the task relies primarily on cross-instance aggregation.

\subsection{Generation-Assisted Understanding Framework}
\label{subsec:gafu_framework}
Our analysis reveals complementary generation-side evidence whose advantage depends on spatial organization and varies across tasks.
GAFU accordingly preserves and aligns this information in compact \(G\) tokens and enables task-conditioned retrieval by the understanding pathway.

\subsubsection{Generation Representation Compression and Alignment}
\paragraph{Compression.}
We adopt an encoder--decoder architecture to compress the generation-side representations into a fixed-size representation \(G\). 
The Encoder concatenates the projected generation-side representations with 256 learnable queries and feeds them into a multi-layer Transformer, where joint self-attention with spatial positional encoding allows the queries to aggregate generation information while retaining its spatial structure. The resulting query outputs are then projected to the same hidden dimension as the understanding-side tokens, forming the \(G\) tokens. 
The Decoder mirrors the Encoder: it concatenates the \(G\) tokens with learnable reconstruction queries and feeds them into a multi-layer Transformer. The resulting query outputs are mapped through a linear prediction head to reconstruct the original generation-side representation. 

\paragraph{Alignment.}
We train this architecture on Cambrian image data, freezing the native understanding and generation pathways of the UMM while optimizing only the Encoder, Decoder, and prediction heads.
The objective consists of reconstruction, spatial binding, and relation alignment losses:
\begin{equation}
\mathcal{L}_{\mathrm{ext}}
=
\lambda_{\mathrm{rec}}\mathcal{L}_{\mathrm{rec}}
+\lambda_{\mathrm{spa}}\mathcal{L}_{\mathrm{spa}}
+\lambda_{\mathrm{rel}}\mathcal{L}_{\mathrm{rel}}.
\label{eq:alignment-objective}
\end{equation}
For reconstruction, the Decoder reconstructs the generation latent solely from \(G\), which is then decoded into an image \(\hat{I}\) through the frozen generation pathway. We define 
\begin{equation}
\mathcal{L}_{\mathrm{rec}}
=
\lVert\hat{I}-I\rVert_1
+0.5\lVert\hat{I}-I\rVert_2
+0.5\,\mathrm{LPIPS}(\hat{I},I).
\label{eq:reconstruction-loss}
\end{equation}
which encourages \(G\) to retain sufficient visual information for reconstructing the original image. 

For spatial binding, we directly map each \(G\) token through a shared linear readout head, without passing it through the Decoder, to predict the spatially aligned generation target:
\begin{equation}
\mathcal{L}_{\mathrm{spa}}
=
\tfrac1N\textstyle\sum_{i=1}^{N}
\left\|h_{\mathrm{spa}}(G_i)-z_i^{\mathrm{bind}}\right\|_2^2.
\label{eq:spatial-binding-loss}
\end{equation}
where \(h_{\mathrm{spa}}\) is the readout head and \(z_i^{\mathrm{bind}}\) is the spatially aligned generation target corresponding to \(G_i\). This explicitly binds \(G\) tokens to the original generation representation's spatial structure.

For relation alignment, we organize \(G\) and the understanding-side representation \(U\) into two-dimensional spatial grids.
For local relation alignment at scale \(s\in\{4,8,16\}\), we obtain \(G^{(s)}=P_s(G)\) and \(U^{(s)}=P_s(U)\) by adaptive average pooling to an \(s\times s\) grid, treat each of the \(s^2\) spatial locations as a node, and compare pairwise relations among these nodes within each image.
For global relation alignment, we mean-pool all spatial locations of each image into a single node and compare pairwise relations among these image-level nodes across the batch.
For a node set \(Z\), we compute its pairwise cosine similarity matrix, remove the diagonal entries, and center the remaining values to form the relation vector \(r(Z)\).
The local and global relation losses are
\begin{equation}
\begin{aligned}
\mathcal{L}_{\mathrm{local}}^{(s)}
&=\tfrac1B\textstyle\sum_{b=1}^{B}
\left[1-\cos\!\left(r(G_b^{(s)}),r(U_b^{(s)})\right)\right],\\
\mathcal{L}_{\mathrm{global}}
&=1-\cos\!\left(r(G^{(1)}),r(U^{(1)})\right).
\end{aligned}
\label{eq:relation-losses}
\end{equation}
where \(G^{(1)}\) and \(U^{(1)}\) contain one spatially mean-pooled node per image, with relations computed across the batch.
The final relation alignment loss is
\begin{equation}
\mathcal{L}_{\mathrm{rel}}
=
\tfrac14\bigl[
\mathcal{L}_{\mathrm{global}}
+\mathcal{L}_{\mathrm{local}}^{(4)}
+\mathcal{L}_{\mathrm{local}}^{(8)}
+\mathcal{L}_{\mathrm{local}}^{(16)}
\bigr].
\label{eq:relation-alignment-loss}
\end{equation}
This aligns the relational structure of \(G\) with the understanding-side representation without directly matching feature values, preserving pathway-specific information while ensuring relational compatibility.
Figure~\ref{fig:gafu-framework} summarizes GAFU, with training configurations provided in Appendix~\ref{app:experimental-details}.

\subsubsection{Task-Conditioned Injection and Fine-Tuning}
We introduce a task-conditioned injection framework and a CE-space pairwise DPO objective~\citep{rafailov2024directpreferenceoptimizationlanguage} to encourage the model to use \(G\) matched with the current sample.

\begin{figure*}[t]
\centering
\includegraphics[width=1.0\textwidth]{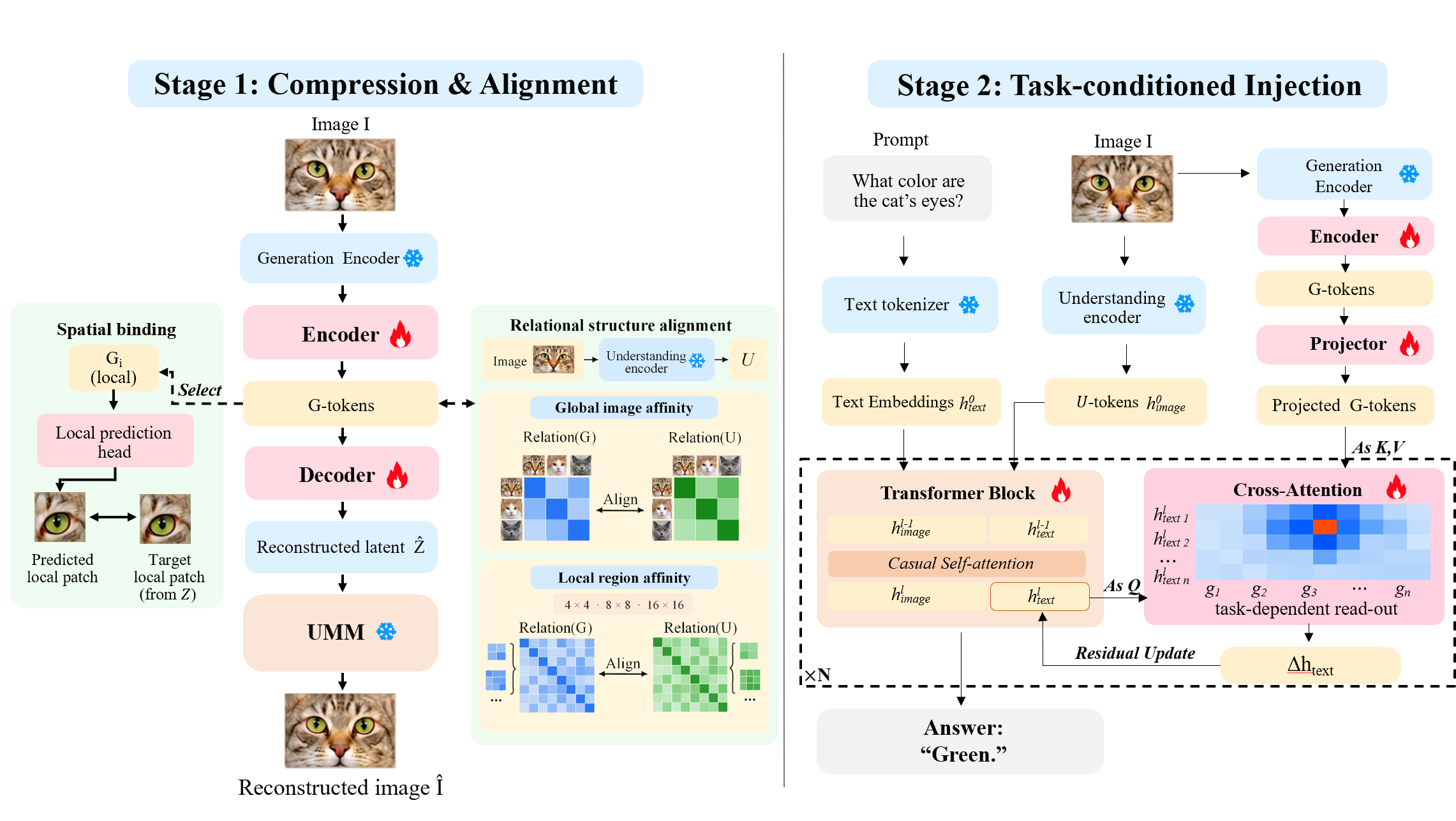}
\caption{Overview of GAFU, comprising generation representation compression and alignment (Stage 1) and task-conditioned injection (Stage 2).}
\label{fig:gafu-framework}
\end{figure*}

\paragraph{Task-Conditioned Injection.}
After each Transformer layer in the understanding pathway, we use current text hidden states as queries and projected \(G\) tokens as keys and values in cross-attention, with \(G\) using the same positional encoding as native understanding-side visual tokens.
The cross-attention output is residually added to the text hidden states and passed to the next layer, allowing selective retrieval of task-relevant information from \(G\) as the textual context evolves.

\paragraph{Training Objective.}
For each training sample \(i\), we construct a positive branch using its matched \(G_i\) and a shuffle-\(G\) branch that replaces \(G_i\) with \(G_{\pi(i)}\) from another image.
We optimize
\begin{equation}
\mathcal{L}
=
\lambda_{\mathrm{CE}}\mathcal{L}_{\mathrm{CE}}
+
\lambda_{\mathrm{DPO}}\mathcal{L}_{\mathrm{DPO}}
+
\lambda_{\mathrm{KL}}(t)\mathcal{L}_{\mathrm{KL}}.
\label{eq:finetuning-objective}
\end{equation}
\textbf{\(\mathcal{L}_{\mathrm{CE}}\)} is the standard answer cross-entropy loss on the positive branch, providing the basic supervision for task learning.
\textbf{\(\mathcal{L}_{\mathrm{DPO}}\)} is a pairwise preference objective defined in CE space, encouraging the positive branch to achieve lower answer CE than the shuffle-\(G\) branch and thereby promoting the use of sample-specific information from \(G\).
\textbf{\(\mathcal{L}_{\mathrm{KL}}\)} is a forward KL loss between the shuffle-\(G\) branch, which receives the correct \(U\) but a mismatched \(G\), and the frozen original model conditioned on \(U\) alone, preserving the original output distribution under mismatched \(G\) and preventing DPO from enlarging the preference gap by degrading the shuffle-\(G\) branch.

Beyond these objectives, we adopt several strategies to stabilize preference learning and encourage use of \(G\).
To stabilize DPO training, we weight positive and negative DPO gradients by \(4:1\) and cap the negative gradient norm at \(0.25\) times that of the matched-\(G\) positive branch, which combines answer-CE and positive DPO gradients, preventing margin growth by degrading the shuffled branch.
The weight of \(\mathcal{L}_{\mathrm{KL}}\) is gradually decayed to zero, stabilizing early preference learning while avoiding over-constraining the model to the \(U\)-only reference later in training.
In early training, we mask \(U\) for a subset of samples to reduce reliance on the original understanding pathway.
Detailed loss definitions, gradient decomposition, and training schedules are provided in Appendix~\ref{app:task-conditioned-finetuning}.

\begin{table}[t]
\centering
\caption{Visual understanding results.}
\label{tab:visual_results}
\resizebox{\linewidth}{!}{%
\begin{tabular}{clccccccc}
\toprule
Type & Model & Size & VStar-Bench$\uparrow$ & OmniSpatial$\uparrow$ & MMVP$\uparrow$ & PathVQA$\uparrow$ & VSR-Bench$\uparrow$ & Avg.$\uparrow$ \\
\midrule

\multirow[c]{4}{*}{\rotatebox[origin=c]{90}{\textit{VLM}}}
& InternVL3 & 1B & 33.51 & 38.81 & 61.67 & 18.74 & 30.65 & 36.67 \\
& InternVL3 & 2B & 41.36 & 36.79 & 67.00 & 28.56 & 41.87 & 43.12 \\
& Qwen3-VL & 2B & 45.03 & 30.99 & 68.33 & 29.94 & 47.65 & 44.39 \\
& Qwen3-VL & 4B & 45.55 & 22.44 & 77.33 & 33.73 & 49.99 & 45.81 \\
\midrule

& BAGEL & 14B & 36.65 & 30.79 & 71.67 & 39.27 & 69.98 & 49.67 \\
& BAGEL (SFT) & 14B & 39.27 & 31.90 & 74.00 & 51.42 & 71.39 & 53.60 \\
\rowcolor{green!12}
\cellcolor{white}
& BAGEL (Ours) & 14B
& \textbf{\underline{47.12}}
& \textbf{\underline{34.51}}
& \textbf{\underline{79.67}}
& \textbf{\underline{59.07}}
& \textbf{\underline{77.54}}
& \textbf{\underline{59.58}} \\
\cmidrule(lr){2-9}

& Janus-Pro & 7B & 31.94 & 25.64 & 54.67 & 16.91 & 55.72 & 36.97 \\
& Janus-Pro (SFT) & 7B & 33.51 & 27.14 & 54.67 & 51.75 & 57.54 & 44.92 \\
\rowcolor{green!12}
\cellcolor{white}
& Janus-Pro (Ours) & 7B
& 38.74
& 31.57
& 59.67
& 52.87
& 59.54
& 48.48 \\
\cmidrule(lr){2-9}

& Lance & 3B & 35.60 & 23.87 & 50.33 & 11.43 & 52.07 & 34.66 \\
& Lance (SFT) & 3B & 36.13 & 24.72 & 50.00 & 43.04 & 53.26 & 41.43 \\
\rowcolor{green!12}
\cellcolor{white}
\multirow[c]{-9}{*}{\raisebox{18pt}{\rotatebox[origin=c]{90}{\textit{UMM}}}}
& Lance (Ours) & 3B
& 42.41
& 29.16
& 53.67
& 46.91
& 54.35
& 45.30 \\
\bottomrule
\end{tabular}%
}
\end{table}

\section{Experiments}
\subsection{Experimental Setup}
\label{subsec:experimental-setup}

\paragraph{Data.}
We train the compression and alignment module on 1M images from Cambrian~\citep{tong2024cambrian1fullyopenvisioncentric}.
For downstream fine-tuning, we use FineVision \citep{wiedmann2026finevisionopendataneed}, Visual Spatial Reasoning (VSR) \citep{liu2023visualspatialreasoning}, and PathVQA \citep{he2020pathvqa30000questionsmedical} for general and domain-specific understanding.
Detailed data composition and preprocessing are provided in Appendix~\ref{app:experimental-details}.

\paragraph{Models and Training.}
We experiment on three UMMs, BAGEL~\citep{deng2025emergingpropertiesunifiedmultimodal}, Janus-Pro~\citep{chen2025janusprounifiedmultimodalunderstanding}, and Lance~\citep{fu2026lanceunifiedmultimodalmodeling}.
During compression and alignment, we freeze the native pathways and optimize the Encoder, Decoder, and prediction heads.
During task-conditioned fine-tuning, we optimize the injection modules and understanding-pathway LoRA adapters, with the compression Encoder later unfrozen for joint optimization.
Training configurations and hyperparameters are provided in
Appendix~\ref{app:experimental-details}.

\paragraph{Evaluation.}
For general fine-grained visual understanding, we evaluate on VStar-Bench, OmniSpatial, and MMVP, covering fine-grained visual perception and spatial understanding.
For domain-specific evaluation, we report results on VSR-Bench and PathVQA.
We compare against the Base model and SFT baselines trained on the same data.
Evaluation details are provided in Appendix~\ref{app:experimental-details}.

\begin{table}[t]
\centering
\caption{Ablation results.}
\label{tab:ablation-results}
\begin{tabular}{llccc}
\toprule
Dimension & Setting & VStar-Bench$\uparrow$ & OmniSpatial$\uparrow$ & MMVP$\uparrow$ \\
\midrule

\multirow[c]{4}{*}{Alignment objective}
& w/o spatial & 46.07 & 33.14 & 75.00 \\
& w/o relation & 45.55 & 34.44 & 74.33 \\
& Reconstruction only & 45.03 & 32.81 & 74.33 \\
& \cellcolor{green!12}Ours (full)
& \cellcolor{green!12}\textbf{\underline{47.12}}
& \cellcolor{green!12}\textbf{\underline{34.51}}
& \cellcolor{green!12}\textbf{\underline{79.67}} \\
\cmidrule(lr){2-5}

\multirow[c]{5}{*}{Injection interface}
& Token-level append & 40.84 & 32.55 & 77.33 \\
& Entry fusion & 41.88 & 22.37 & 75.00 \\
& $q=h_{\mathrm{img}}$ & 43.98 & 33.46 & 75.00 \\
& $q=\text{Learnable query}$ & 42.41 & 34.18 & 75.67 \\
& \cellcolor{green!12}Ours ($q=h_{\mathrm{text}}$)
& \cellcolor{green!12}\textbf{\underline{47.12}}
& \cellcolor{green!12}\textbf{\underline{34.51}}
& \cellcolor{green!12}\textbf{\underline{79.67}} \\
\cmidrule(lr){2-5}

\multirow[c]{2}{*}{Injection density}
& Sparse-layer & 46.60 & 34.51 & 78.00 \\
& \cellcolor{green!12}Ours (full-layer)
& \cellcolor{green!12}\textbf{\underline{47.12}}
& \cellcolor{green!12}\textbf{\underline{34.51}}
& \cellcolor{green!12}\textbf{\underline{79.67}} \\
\cmidrule(lr){2-5}

\multirow[c]{3}{*}{Fine-tuning loss}
& CE only & 42.93 & 33.33 & 65.33 \\
& CE + DPO & 43.98 & 22.24 & 73.33 \\
& \cellcolor{green!12}Ours (CE + DPO + KL)
& \cellcolor{green!12}\textbf{\underline{47.12}}
& \cellcolor{green!12}\textbf{\underline{34.51}}
& \cellcolor{green!12}\textbf{\underline{79.67}} \\
\bottomrule
\end{tabular}
\end{table}

\subsection{General Fine-grained Visual Understanding}
We evaluate whether GAFU provides consistent gains beyond SFT for general fine-grained visual understanding across UMM architectures.
As shown in Table~\ref{tab:visual_results}, GAFU outperforms SFT on all three benchmarks across BAGEL, Janus-Pro, and Lance, improving VStar-Bench by 5.23--7.85 percentage points, OmniSpatial by 2.61--4.44 points, and MMVP by 3.67--5.67 points.
The consistent gains from 3B Lance to 14B BAGEL show that generation-side complementarity benefits fine-grained visual understanding across model scales, UMM architectures, and representation pathways.

Across all nine model--benchmark combinations, the improvement of our method over SFT is larger than that of SFT over the Base model. This contrast is particularly evident on MMVP, where SFT brings no improvement to Janus-Pro and slightly degrades Lance, whereas our method achieves further gains, suggesting that supervised adaptation relying solely on the native understanding pathway cannot reliably overcome its fine-grained visual representation bottleneck, while complementary generation-side information provides additional effective visual evidence.
The largest gains on VStar-Bench indicate that generation-side representations provide complementary evidence for local attributes, small objects, and relative positions that can be weakened by visual compression in the understanding pathway. Consistent OmniSpatial gains show that this complementarity extends to spatial layouts and inter-object relations, enabling finer spatial reasoning and showing that this advantage spans local details and spatial relations rather than one type of visual evidence.

\subsection{Domain-Specific Visual Understanding}

We further examine whether generation-side information remains beneficial after domain-specific adaptation.
As shown in Table~\ref{tab:visual_results}, our method outperforms SFT on both tasks across all three models, improving PathVQA by 1.12--7.65 percentage points and VSR-Bench by 1.09--6.15 points.
These consistent post-adaptation gains show that domain supervision does not fully capture the complementary visual evidence available in generation-side representations.

\begin{figure*}[t]
    \centering
    \includegraphics[width=\textwidth]{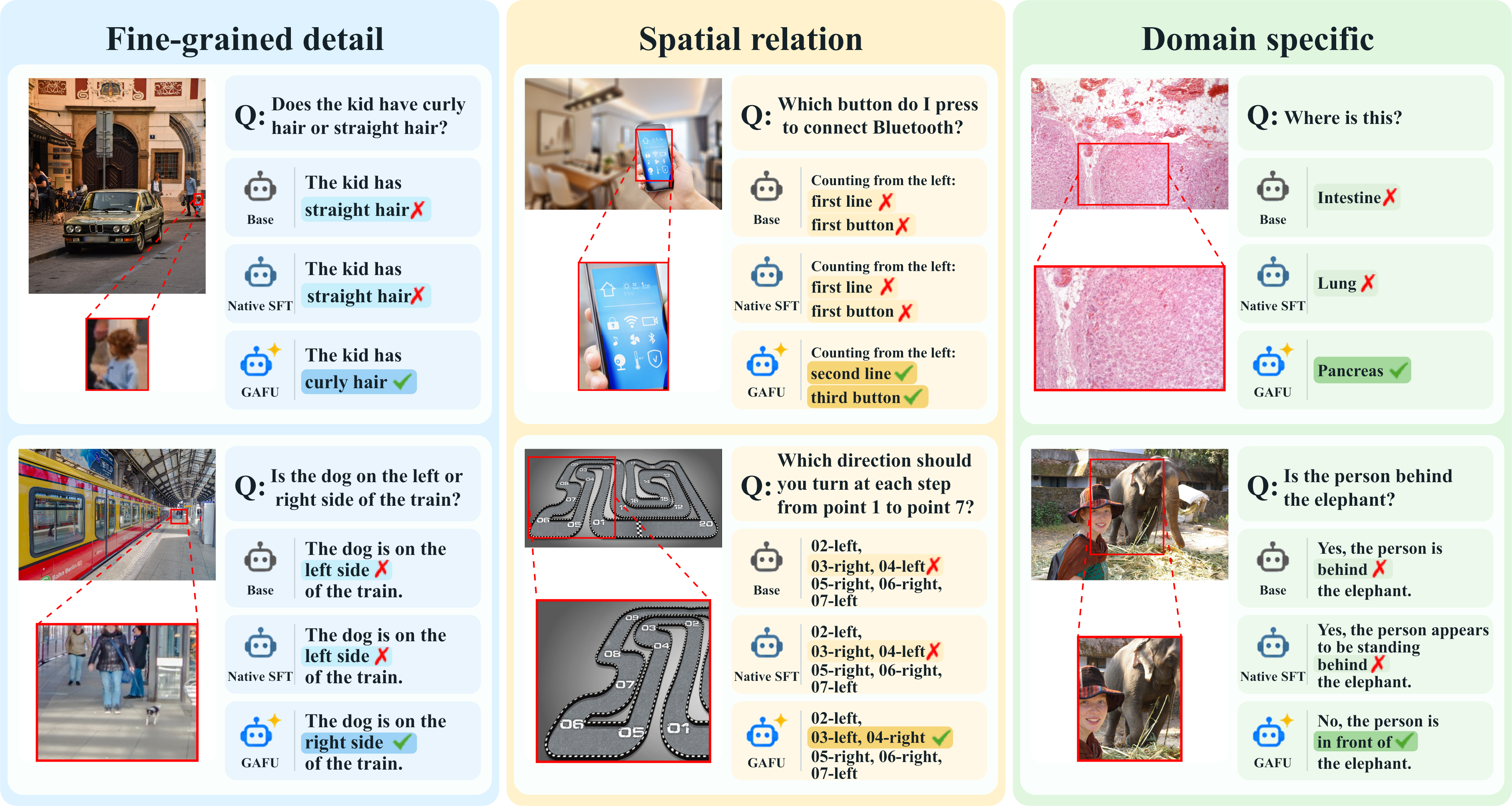}
    \caption{
    \textbf{GAFU enhances fine-grained visual understanding.}
    Examples show better recognition of fine-grained details, spatial relations,
    and domain-specific cues over the base model and SFT.
    }
    \label{fig:teaser}
\end{figure*}

The two domains show distinct gain patterns, reflecting different roles of generation-side information.
On PathVQA, SFT already provides substantial gains over the Base models, yet our method further improves upon these strong SFT baselines, indicating that although domain supervision enables substantial medical-domain and task adaptation, it does not fully resolve the use of fine-grained visual evidence, which can be further complemented by generation-side information.
In contrast, VSR-Bench shows a sharper pattern: SFT yields only marginal gains over Base, whereas GAFU provides clear additional gains, indicating that the remaining limitation is more strongly tied to spatial and relational representation than to task adaptation. The structural and spatial information retained by generation-side representations directly addresses this limitation.
Representative qualitative examples across general and domain-specific tasks are shown in Figure~\ref{fig:teaser}.

\subsection{Ablation Study}
We conduct all ablations on BAGEL under the general setting, covering alignment objectives, injection interfaces, injection density, and fine-tuning loss.
For alignment objectives, we remove spatial binding and relation alignment separately or together, with reconstruction unchanged.
For injection interfaces, we compare concatenating \(G\) with the native understanding-side visual tokens \(U\), querying \(G\) with \(U\), current visual hidden states, or input-independent learnable queries. Our method queries \(G\) with current text hidden states.
For injection density, we compare all 27 Transformer layers with 10 evenly spaced layers.
For fine-tuning loss, we compare CE only, CE + DPO, and the full CE + DPO + KL objective.
Results are reported in Table~\ref{tab:ablation-results}.


\textbf{Alignment Objective.}
OmniSpatial highlights the role of spatial binding in preserving spatial correspondences, dropping by 1.37 points without it, while VStar-Bench shows that fine-grained evidence requires relation alignment with the understanding pathway, dropping by 1.57 points without it.
MMVP further reveals their complementarity: combining both constraints outperforms either by 4.67--5.34 points, showing that their joint use is key to exploiting generation-side information.

\textbf{Injection Interface.}
How \(G\) is read matters more than merely making it available to the model.
Concatenating \(G\) with \(U\), or using \(U\) to query \(G\) at input, trails ours by 1.96--6.28 and 4.67--12.14 points, respectively, suggesting that premature fusion can interfere with understanding representations before the language-side task state is formed.
Visual hidden-state queries outperform input-level fusion but remain 1.05--4.67 points below text hidden-state queries across benchmarks. Learnable queries nearly match ours on OmniSpatial with a 0.33-point gap but lag by 4.71 and 4.00 points on VStar-Bench and MMVP, showing that task-independent queries can capture stable spatial structure, whereas selecting question-specific evidence requires task context from text hidden states.

\textbf{Injection Density.}
10-layer sparse injection remains within 0.00--1.67 points of full-layer performance, showing that injection frequency matters less than how \(G\) is organized and accessed.

\textbf{Fine-Tuning Loss.}
DPO improves VStar-Bench and MMVP by 1.05 and 8.00 points but degrades OmniSpatial by 11.09 points without KL. The full objective then improves all three benchmarks by 3.14--12.27 points over CE + DPO, showing that DPO promotes sample-specific \(G\) use while KL is essential for turning preference learning into reliable gains.

\section{Conclusion}
We present GAFU, a generation-assisted understanding post-training framework for fine-grained visual understanding in UMMs.
First, we systematically examine the role of native generation-side
representations in fine-grained understanding and the conditions that enable it.
Second, based on these findings, GAFU preserves and organizes complementary generation-side information through compression, spatial binding, and relation alignment, selectively retrieves it through task-conditioned injection, and promotes the use of sample-relevant evidence through pairwise preference learning.
Third, across UMM architectures, GAFU consistently outperforms same-data SFT on general, spatial, and domain-specific visual understanding tasks, with maximum relative gains of 20.00\%, 17.94\%, and 8.62\% on VStar-Bench, OmniSpatial, and VSR-Bench, respectively.
Overall, GAFU converts complementary information in native generation-side representations into usable visual evidence for the understanding pathway, yielding consistent gains in fine-grained visual understanding.
We hope this work will help UMMs move beyond joint modeling of understanding and generation toward deeper capability synergy, and provide a new path toward more reliable and fine-grained visual understanding in complex real-world scenarios.

\section*{AI Use Statement}

Generative AI tools were used to assist with translation and to edit the
research paper to improve readability. The tools were not used to generate
synthetic data sets; help develop theoretical models or conceptual frameworks;
formulate mathematical claims; provide critical ingredients for proving
mathematical claims; assist in the writing of proofs; propose or refine
hypotheses; design or provide feedback on research methodology or experiments;
implement methods; clean or reformat datasets; support qualitative or thematic
data analysis; interpret results; formulate questions for surveys or
interviews; create or modify scientific figures or images; suggest
experimental parameters; create or edit software code; create artifacts; draft
parts of the research paper; transcribe recordings of research material;
summarize or analyse existing literature; discover research topics or identify
research gaps; brainstorm; source or search for information; identify relevant
literature; format references; suggest a structure for the research paper; or
propose a title or keywords. All AI-assisted text was reviewed by the authors,
who take full responsibility for the final content of this submission.

\section*{Ethics Statement}

This work uses existing datasets and benchmark suites, including Cambrian,
FineVision, VStar-Bench, OmniSpatial, MMVP, VSR-Bench, VisMin, and PathVQA, for model
training and evaluation, in accordance with their respective data-use and
citation requirements. The medical question-answering results on PathVQA are
intended solely for research benchmarking and should not be used to support
clinical decisions. This work does not involve privacy-sensitive data, human
subjects, new data collection, dataset release, or the development of
potentially harmful insights. The authors declare no conflicts of interest or
external sponsorship related to this work.

\section*{Reproducibility statement}
For reproducing the systematic analysis:
Section~\ref{subsec:systematic-analysis} introduces the overall setup, research questions, and analysis procedure. Appendix~\ref{app:representation-sensitivity}
further specifies the extraction and preprocessing of understanding- and
generation-side representations across UMMs, together with the computation of the sensitivity, resolution, and spatial-dependence analyses.

For reproducing GAFU training and evaluation:
Section~\ref{subsec:gafu_framework} describes the overall GAFU architecture
and the basic procedures of Stage~1 and Stage~2, while
Section~\ref{subsec:experimental-setup} summarizes the tasks, models, and
overall experimental setup. Appendix~\ref{app:compression-alignment} specifies
the Stage~1 GToken Encoder--Decoder architecture and the computation of the
reconstruction, spatial-binding, and relation-alignment objectives.
Appendix~\ref{app:task-conditioned-finetuning} describes the Stage~2 injection
modules, training objectives, gradient decomposition, masking schedule, and shuffle-donor construction. Appendix~\ref{app:experimental-details}
provides the data composition, preprocessing examples, optimizer and training
configurations, and evaluation protocols.

\bibliography{main}
\bibliographystyle{iclr2027_conference}

\newpage
\appendix

\section{Details of Representation Sensitivity Analysis}
\label{app:representation-sensitivity}

\subsection{Representation Extraction and Preprocessing}
\label{app:representation-extraction}

We extract the understanding-side representation \(R_U\) and generation-side representation \(R_G\) from each UMM as follows:
\begin{itemize}
    \item \textbf{BAGEL.} \(R_U\) comprises the SigLIP visual tokens after the connector (\(16\times16\)). \(R_G\) is obtained from the KL-VAE posterior mean (\(64\times64\)) through \(2\times2\) patchification, yielding a \(32\times32\) grid.

    \item \textbf{Janus-Pro.} \(R_U\) comprises the SigLIP-L/16 visual tokens after the aligner, and \(R_G\) comprises the quantized VQ-16 codebook embeddings \(Q_{\mathrm{vq}}\), both on \(24\times24\) grids.

    \item \textbf{Lance.} \(R_U\) comprises the merged Qwen visual tokens (\(16\times16\)), and \(R_G\) is the posterior mean of the frozen Wan2.2-VAE (\(32\times32\)).

    \item \textbf{Show-o2-7B.} \(R_U\) is extracted from the output of
    \texttt{image\_embedder\_und} after adding the native visual positional
    embedding and applying \texttt{und\_trans}. \(R_G\) is extracted from the
    output of \texttt{image\_embedder\_gen}. Both representations are
    adaptively average-pooled to a common \(16\times16\) grid. The resulting
    channel dimensions are 1152 for \(R_U\) and 3584 for \(R_G\).

    \item \textbf{BLIP3-o-4B.} \(R_U\) comprises the merged Qwen2.5-VL visual tokens (\(16\times16\)), and \(R_G\) comprises the penultimate-layer patch features of the EVA-CLIP-E/14-plus generation vision tower (\(32\times32\)).

    \item \textbf{DreamLLM-7B.} \(R_U\) comprises the penultimate-layer CLIP ViT-L/14 patch features after the native projector, excluding the CLS token (\(16\times16\)). \(R_G\) is the generation VAE posterior mean with the native latent scaling factor applied (\(64\times64\)).
\end{itemize}

We preserve the spatial arrangement of each representation and align its grid to \(16\times16\) using adaptive average pooling, yielding the position-wise features \(\bar{R}(I,p)\) used in the main text. For VAE-based representations, each non-overlapping \(2\times2\) latent
block is concatenated along the channel dimension to form one patch feature before adaptive pooling. Representations that are already provided as patch-level features, such as Show-o2-7B and BLIP3-o-4B, are pooled directly from their native feature grids. This procedure standardizes the spatial grid without requiring \(R_U\) and \(R_G\) to have the same feature dimension.

For the sensitivity metric, cosine similarity uses \(\epsilon_{\mathrm{cos}}=10^{-6}\). Local responses are averaged within contiguous \(2\times2\) windows with stride one and no padding, followed by maximum aggregation.

\subsection{Control-Corrected Construction}

Our goal is to construct a control image \(I_n\) for each semantic edit \(I\rightarrow I_v\) that has the same visual change intensity within the same edited region, while preserving natural appearance and subject semantics.

We first identify the edited region from the pixel differences between \(I\) and \(I_v\). For each pixel \(p\), we compute the mean squared RGB difference:

\[
e(p)=\frac{1}{3}\sum_{c=1}^{3}
\bigl(I_v(p,c)-I(p,c)\bigr)^2.
\]

We select the 25\% of nonzero pixels with the largest differences and apply two binary dilations to obtain the edited-region mask \(M\). All control changes are confined to \(M\), while pixels outside \(M\) remain identical to the original image.

We convert the images to linear RGB and denote the original and semantic-edit images by \(X\) and \(X_v\). The RMS within the edited region is defined as:

\[
\operatorname{RMS}_{M}(A)
=
\sqrt{
\frac{1}{3|M|}
\sum_{p\in M}\sum_{c=1}^{3}A(p,c)^2
}.
\]

We generate candidates using predefined single structural-preserving natural degradation operators, including white- or black-endpoint mixing, gamma curves, local-mean smoothing, regional-mean replacement, horizontal or vertical illumination redistribution, spatial illumination fields, piecewise tone curves, luminance lifting, and local-contrast adjustment. These operators do not translate, scale, deform, add, remove, or redraw the subject.

For a valid endpoint \(J\) of an operator, we define its local direction as:

\[
D=w\odot(J-X),
\qquad
w(p)=M(p)\min\left(\frac{\operatorname{dist}(p,\partial M)}{8},1\right).
\]

The control image is then solved in closed form along this single direction:

\[
X_n
=
X+
\frac{\operatorname{RMS}_{M}(X_v-X)}
{\operatorname{RMS}_{M}(D)}
D.
\]

Thus, \(X_n\) and \(X_v\) have the same linear-RGB RMS within the edited region. We require \(X_n\in[0,1]\) and discard candidates that violate this constraint. Each operator produces one candidate; the remaining candidate closest to the original image under LPIPS is selected as the final control image.

\subsection{Complementarity Analysis}

Using the strong-response criterion defined in the main text,
Table~\ref{tab:complementarity} reports the proportions of samples with
generation-side-only and understanding-side-only strong responses, together
with their difference.
Across all 24 model--domain combinations, the proportion of generation-side-only
strong responses is higher than that of understanding-side-only strong responses.
The average values of
\(P_{G\text{-only}}\) and \(P_{U\text{-only}}\) are \(28.18\%\) and \(2.48\%\),
respectively, and the average net difference is \(25.70\) percentage points.
This result indicates that generation-side representations exhibit unique
strong responses to a broad range of fine-grained visual changes.

Across domains, attribute and object have the highest
\(P_{G\text{-only}}\) values, at \(35.66\%\) and \(33.19\%\), respectively.
This indicates that generation-side-only responses occur more frequently for
attribute- and object-related changes. The generation-side advantages in the
count and relation domains are relatively weaker, with values of \(23.29\%\)
and \(20.58\%\), respectively, consistent with the smaller \(G-U\) gaps in
these two domains observed in the sensitivity experiments.

\begin{table}[t]
\centering
\caption{Strong-response complementarity between generation-side and understanding-side representations.}
\label{tab:complementarity}
\begin{tabular}{llrrr}
\toprule
Model & Domain & $P_{G\text{-only}}^{(d)}$ & $P_{U\text{-only}}^{(d)}$ & $\Delta_{\mathrm{comp}}^{(d)}$ \\
\midrule
BAGEL & \texttt{attribute} & 32.99\% & 2.04\% & $+30.95$ pp \\
BAGEL & \texttt{object} & 28.37\% & 2.94\% & $+25.43$ pp \\
BAGEL & \texttt{count} & 9.04\% & 7.12\% & $+1.92$ pp \\
BAGEL & \texttt{relation} & 19.26\% & 0.84\% & $+18.43$ pp \\
Janus-Pro & \texttt{attribute} & 51.02\% & 1.36\% & $+49.66$ pp \\
Janus-Pro & \texttt{object} & 48.27\% & 0.52\% & $+47.75$ pp \\
Janus-Pro & \texttt{count} & 28.22\% & 3.84\% & $+24.38$ pp \\
Janus-Pro & \texttt{relation} & 23.45\% & 0.67\% & $+22.78$ pp \\
Lance & \texttt{attribute} & 57.14\% & 0.34\% & $+56.80$ pp \\
Lance & \texttt{object} & 52.94\% & 0.17\% & $+52.77$ pp \\
Lance & \texttt{count} & 29.32\% & 2.74\% & $+26.58$ pp \\
Lance & \texttt{relation} & 34.51\% & 0.00\% & $+34.51$ pp \\
Show-o2-7B & \texttt{attribute} & 26.87\% & 8.50\% & $+18.37$ pp \\
Show-o2-7B & \texttt{object} & 22.32\% & 8.13\% & $+14.19$ pp \\
Show-o2-7B & \texttt{count} & 43.01\% & 0.82\% & $+42.19$ pp \\
Show-o2-7B & \texttt{relation} & 25.29\% & 4.86\% & $+20.44$ pp \\
BLIP3-o-4B & \texttt{attribute} & 21.09\% & 0.00\% & $+21.09$ pp \\
BLIP3-o-4B & \texttt{object} & 25.26\% & 0.69\% & $+24.57$ pp \\
BLIP3-o-4B & \texttt{count} & 10.68\% & 5.75\% & $+4.93$ pp \\
BLIP3-o-4B & \texttt{relation} & 7.37\% & 0.00\% & $+7.37$ pp \\
DreamLLM-7B & \texttt{attribute} & 24.83\% & 0.00\% & $+24.83$ pp \\
DreamLLM-7B & \texttt{object} & 21.97\% & 0.87\% & $+21.11$ pp \\
DreamLLM-7B & \texttt{count} & 19.45\% & 7.12\% & $+12.33$ pp \\
DreamLLM-7B & \texttt{relation} & 13.57\% & 0.17\% & $+13.40$ pp \\
\bottomrule
\end{tabular}
\end{table}
\subsection{Resolution Stability}

To test whether the generation-side advantage depends on input scale, we repeat the same image preprocessing, representation extraction, common-grid alignment, and control-correction procedures at input resolutions of 128, 256, and 512. At each resolution, the long edge is first reduced to the target resolution and then restored to the model input size by interpolation.
We report the control-corrected generation--understanding response gap for all six models, four domains, and three input resolutions in Table~\ref{tab:resolution-stability}.
\begin{table}[t]
\centering
\caption{Resolution stability of the control-corrected generation--understanding response gap.}
\label{tab:resolution-stability}
\begin{tabular}{llrrr}
\toprule
Model & Domain & 128 & 256 & 512 \\
\midrule
BAGEL & \texttt{attribute} & 0.391 & 0.334 & 0.468 \\
BAGEL & \texttt{object} & 0.158 & 0.149 & 0.303 \\
BAGEL & \texttt{count} & 0.063 & 0.102 & 0.148 \\
BAGEL & \texttt{relation} & 0.221 & 0.183 & 0.238 \\
Janus-Pro & \texttt{attribute} & 0.787 & 0.783 & 0.803 \\
Janus-Pro & \texttt{object} & 0.682 & 0.735 & 0.727 \\
Janus-Pro & \texttt{count} & 0.067 & 0.202 & 0.330 \\
Janus-Pro & \texttt{relation} & 0.512 & 0.492 & 0.489 \\
Lance & \texttt{attribute} & 0.830 & 0.929 & 1.095 \\
Lance & \texttt{object} & 0.719 & 0.809 & 1.004 \\
Lance & \texttt{count} & 0.498 & 0.453 & 0.469 \\
Lance & \texttt{relation} & 0.496 & 0.559 & 0.675 \\
Show-o2-7B & \texttt{attribute} & 0.549 & 0.526 & 0.587 \\
Show-o2-7B & \texttt{object} & 0.250 & 0.340 & 0.457 \\
Show-o2-7B & \texttt{count} & 0.583 & 0.715 & 0.875 \\
Show-o2-7B & \texttt{relation} & 0.274 & 0.311 & 0.376 \\
BLIP3-o-4B & \texttt{attribute} & 0.379 & 0.334 & 0.376 \\
BLIP3-o-4B & \texttt{object} & 0.394 & 0.363 & 0.393 \\
BLIP3-o-4B & \texttt{count} & 0.211 & 0.072 & 0.092 \\
BLIP3-o-4B & \texttt{relation} & 0.267 & 0.265 & 0.294 \\
DreamLLM-7B & \texttt{attribute} & 0.501 & 0.288 & 0.343 \\
DreamLLM-7B & \texttt{object} & 0.283 & 0.150 & 0.235 \\
DreamLLM-7B & \texttt{count} & 0.019 & 0.053 & 0.089 \\
DreamLLM-7B & \texttt{relation} & 0.327 & 0.225 & 0.225 \\
\bottomrule
\end{tabular}
\end{table}

The corrected generation--understanding response gap remains positive for all six models across all domains and resolutions, indicating a stable generation-side advantage.
\subsection{Spatial Dependence}
To test whether the generation-side response depends on the original spatial organization, we apply a fixed block-level spatial permutation before encoding. 
Because the six models use different input canvases, permutation is
performed on a \(512\times512\) canvas for BAGEL, Lance, and DreamLLM-7B,
on a \(384\times384\) canvas for Janus-Pro and Show-o2-7B, and on a
\(448\times448\) canvas for BLIP3-o-4B. In all cases, the canvas is
partitioned into a fixed \(64\times64\) grid of image blocks, corresponding
to \(8\times8\), \(6\times6\), and \(7\times7\)-pixel blocks, respectively.
The permuted and unpermuted conditions use identical representation
extraction, common-grid alignment, and control-correction procedures.

Table~\ref{tab:spatial-dependence} reports the control-corrected responses before and after spatial permutation.

\begin{table*}[t]
\centering
\small
\setlength{\tabcolsep}{3pt}
\caption{Spatial dependence of control-corrected representation sensitivity.}
\label{tab:spatial-dependence}
\begin{tabular}{llrrrrrr}
\toprule
Model & Domain & $G$ & $G_{\mathrm{shuffle}}$ & $U$ & $U_{\mathrm{shuffle}}$ & $G-U$ & $(G-U)_{\mathrm{shuffle}}$ \\
\midrule
BAGEL & \texttt{attribute} & 0.764 & 0.454 & 0.297 & 0.339 & 0.468 & 0.116 \\
BAGEL & \texttt{object} & 0.667 & 0.402 & 0.364 & 0.296 & 0.303 & 0.106 \\
BAGEL & \texttt{count} & 0.473 & 0.466 & 0.325 & 0.380 & 0.148 & 0.086 \\
BAGEL & \texttt{relation} & 0.418 & 0.045 & 0.180 & 0.173 & 0.238 & -0.128 \\
Janus-Pro & \texttt{attribute} & 1.095 & 0.264 & 0.293 & 0.157 & 0.803 & 0.107 \\
Janus-Pro & \texttt{object} & 1.088 & 0.236 & 0.361 & 0.172 & 0.727 & 0.063 \\
Janus-Pro & \texttt{count} & 1.006 & 0.232 & 0.676 & 0.296 & 0.330 & -0.064 \\
Janus-Pro & \texttt{relation} & 0.711 & 0.197 & 0.222 & 0.070 & 0.489 & 0.127 \\
Lance & \texttt{attribute} & 1.349 & 1.108 & 0.255 & 0.188 & 1.095 & 0.920 \\
Lance & \texttt{object} & 1.298 & 1.067 & 0.294 & 0.173 & 1.004 & 0.894 \\
Lance & \texttt{count} & 1.165 & 0.639 & 0.696 & 0.193 & 0.469 & 0.446 \\
Lance & \texttt{relation} & 0.792 & 0.521 & 0.117 & 0.050 & 0.675 & 0.472 \\
Show-o2-7B & \texttt{attribute} & 1.518 & 1.127 & 0.932 & 0.710 & 0.585 & 0.417 \\
Show-o2-7B & \texttt{object} & 1.490 & 1.051 & 1.033 & 0.669 & 0.457 & 0.382 \\
Show-o2-7B & \texttt{count} & 1.680 & 0.899 & 0.806 & 0.477 & 0.874 & 0.422 \\
Show-o2-7B & \texttt{relation} & 0.965 & 0.601 & 0.588 & 0.409 & 0.377 & 0.192 \\
BLIP3-o-4B & \texttt{attribute} & 0.626 & 0.388 & 0.250 & 0.189 & 0.376 & 0.199 \\
BLIP3-o-4B & \texttt{object} & 0.687 & 0.388 & 0.293 & 0.182 & 0.393 & 0.206 \\
BLIP3-o-4B & \texttt{count} & 0.748 & 0.306 & 0.656 & 0.228 & 0.092 & 0.078 \\
BLIP3-o-4B & \texttt{relation} & 0.406 & 0.121 & 0.112 & 0.062 & 0.294 & 0.059 \\
DreamLLM-7B & \texttt{attribute} & 0.544 & 0.222 & 0.201 & 0.173 & 0.343 & 0.049 \\
DreamLLM-7B & \texttt{object} & 0.457 & 0.183 & 0.222 & 0.163 & 0.235 & 0.020 \\
DreamLLM-7B & \texttt{count} & 0.673 & 0.245 & 0.584 & 0.166 & 0.089 & 0.079 \\
DreamLLM-7B & \texttt{relation} & 0.348 & -0.166 & 0.123 & 0.085 & 0.225 & -0.251 \\
\bottomrule
\end{tabular}
\end{table*}

Spatial permutation reduces the generation--understanding response gap in all 24 model--domain combinations, supporting the dependence of the generation-side advantage on spatial organization.

\section{Compression and Alignment}
\label{app:compression-alignment}

In Stage 1, we freeze the native understanding and generation pathways of the unified multimodal model and optimize only the GToken Encoder, Decoder, and their prediction heads.

The GToken Encoder--Decoder adopts a symmetric six-layer Transformer architecture with a hidden size of 1024, 8 attention heads, and a 4096-dimensional FFN, together with 2D RoPE for spatial position encoding. The Encoder first projects the generation representation to 1024 dimensions and concatenates it with 256 learnable queries. Joint self-attention with spatial positional encoding then aggregates the generation information. The query outputs are subsequently projected to the hidden dimension of the corresponding understanding-pathway tokens, forming a fixed-length generation memory \(G\). The dimensions of \(G\) for BAGEL, Janus-Pro, and Lance are 3584, 4096, and 2048, respectively. The Decoder projects \(G\) back to 1024 dimensions, concatenates it with reconstruction queries arranged according to the source grid, and reconstructs the original generation representation through joint self-attention and a linear prediction head.

For spatial binding, we construct \(z_i^{\mathrm{bind}}\) on the
\(16\times16\) grid of the 256 \(G\) tokens. For BAGEL and Lance,
\(z_i^{\mathrm{bind}}\) is formed by concatenating the corresponding
\(2\times2\) region of the \(32\times32\) grid obtained from the native
generation latent. For Janus-Pro, it is obtained by adaptively average-pooling
the native \(24\times24\) grid to \(16\times16\). Each \(G_i\) predicts the
\(z_i^{\mathrm{bind}}\) at the same spatial location, using a target-energy-
normalized mean squared error.

\begin{figure*}[t]
    \centering
    \includegraphics[
        width=\textwidth
    ]{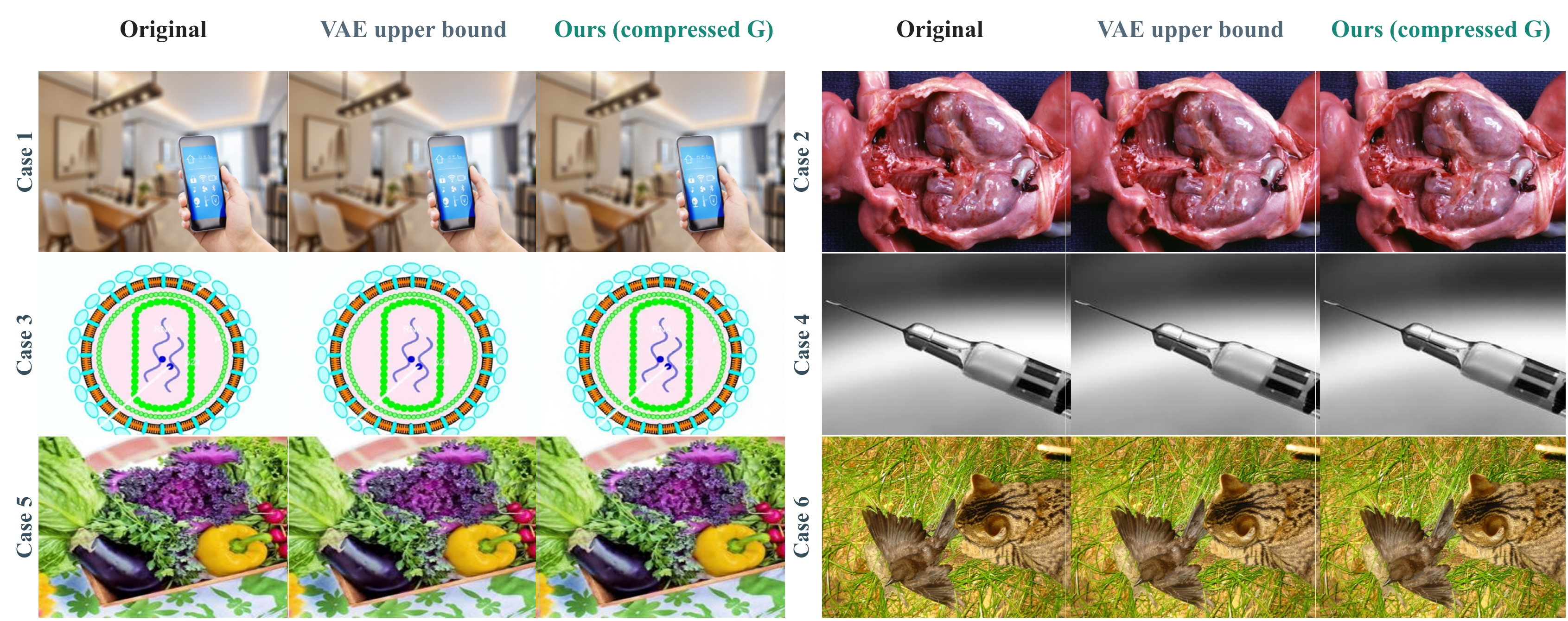}
    \caption{
    \textbf{Stage 1 reconstruction visualization.}
    Each case compares the original image, the VAE upper bound obtained by decoding the native generation representation, and the reconstruction from the compressed \(G\) tokens.
    }
    \label{fig:stage1-reconstruction}
\end{figure*}

Figure~\ref{fig:stage1-reconstruction} visualizes reconstructions on held-out images. Each case compares the original image, the reconstruction obtained by directly decoding the native generation-side representation (VAE upper bound), and the reconstruction obtained from the compressed \(G\) tokens.

\section{Task-Conditioned Injection and Fine-Tuning}
\label{app:task-conditioned-finetuning}

In the initial stage of Stage 2, we freeze the native visual encoders, the Stage 1 GToken Encoder--Decoder, and the LLM backbone, and optimize only the projector, the internal cross-attention injection modules, and the LoRA modules in the understanding pathway. During the latter half of training, we unfreeze the GToken Encoder and jointly optimize it with the projector, internal injection modules, and LoRA modules. The native visual encoders remain frozen throughout training.

The projector maps the \(G\) tokens to the corresponding understanding width and uses the same positional encoding scheme as the native understanding-side visual tokens.

Each internal attention injection block uses the current-layer text hidden states in the understanding pathway as queries and projects the queries, keys, and values into a 1024-dimensional, 16-head attention space. The retrieved features are passed through an output projection and residually written back to the text hidden states. The output projection is zero-initialized so that the injection modules do not alter the original understanding pathway at the beginning of training. The LoRA rank and alpha are both set to 16.

The overall training objective consists of answer supervision on matched samples, a CE-space pairwise DPO objective between matched and mismatched generation memories, and a KL regularizer on the shuffle-\(G\) branch:

\[
\mathcal{L}
=
\lambda_{\mathrm{CE}}\mathcal{L}_{\mathrm{CE}}^{\mathrm{real}}
+
\lambda_{\mathrm{DPO}}\mathcal{L}_{\mathrm{DPO}}
+
\lambda_{\mathrm{KL}}(t)
\mathcal{L}_{\mathrm{KL}}^{\mathrm{shuffle}}.
\]

Here, \(\mathcal{L}_{\mathrm{CE}}^{\mathrm{real}}\) provides answer supervision with the matched \(G\) tokens; \(\mathcal{L}_{\mathrm{DPO}}\) compares answer CE under matched and mismatched generation memories to encourage the use of sample-relevant generation information; and \(\mathcal{L}_{\mathrm{KL}}^{\mathrm{shuffle}}\) constrains the effect of mismatched \(G\) tokens on the original output distribution. In the default configuration, \(\lambda_{\mathrm{CE}}=5\), \(\lambda_{\mathrm{DPO}}=1\), and the DPO temperature is \(\beta=0.1\). The positive and negative DPO branches are assigned gradient weights \(w_{+}=4\) and \(w_{-}=1\), respectively, and the norm of the negative gradient is capped at \(0.25\) times the norm of the positive gradient. The base weight of the shuffle-\(G\) KL loss is \(1.0\); its effective coefficient \(\lambda_{\mathrm{KL}}(t)\) remains at \(1\) initially, then linearly decays to \(0.25\), briefly remains at \(0.25\), and finally linearly decays to \(0\) over the 5,000 optimizer steps.

For each training sample \(i\), let \((q_i,y_i)\) denote the question and answer, \(U_i\) the native understanding representation, and \(G_i\) the corresponding \(G\) tokens. We construct the following three branches:

\begin{itemize}
\item The real/positive branch uses \((q_i,U_i,G_i)\), where \(G_i\) is matched to the current sample, and computes answer-token CE under teacher forcing with the answer \(y_i\).

\item The shuffle-\(G\) branch uses \((q_i,U_i,G_{\pi(i)})\), replacing only the generation memory with a donor memory while retaining the current sample's question, answer, and understanding representation. Donors are selected deterministically by a half-manifest offset: for a manifest containing an even number \(N\) of samples, \(\pi(i)=(i+N/2)\bmod N\).


\item The reference branch does not receive the \(G\) tokens and disables the generation-memory injection modules and understanding-pathway LoRA. It uses the frozen native model to compute the reference distribution conditioned on \((q_i,U_i)\):
\[
p_{\mathrm{ref}}(\cdot\mid q_i,U_i,y_{i,<t}).
\]
This branch does not participate in backpropagation.
\end{itemize}

\paragraph{CE loss.}

Let \(\mathcal{T}_i\) denote the set of answer-token positions for sample \(i\), and let \(n_i=|\mathcal{T}_i|\). In the real/positive branch, the model receives the matched \(G_i\) and computes the negative log-likelihood of the correct answer tokens:

\[
s_i^+
=
-\sum_{t\in\mathcal{T}_i}
\log p_{\theta}
\left(
y_{i,t}
\mid q_i,U_i,G_i,y_{i,<t}
\right),
\]

\[
\ell_i^+
=
\frac{s_i^+}{n_i}.
\]

The real-pair CE loss is defined as:

\[
\mathcal{L}_{\mathrm{CE}}^{\mathrm{real}}
=
\frac{1}{B}
\sum_{i=1}^{B}
\ell_i^+.
\]

\paragraph{CE-space pairwise DPO objective.}

To encourage the model to use the \(G\) tokens matched to the current sample, we replace \(G_i\) with the \(G\) tokens \(G_{\pi(i)}\) from another image to construct the shuffle-\(G\) branch. This branch keeps \(q_i\), \(y_i\), and \(U_i\) unchanged and replaces only the \(G\) tokens:

\[
s_i^-
=
-\sum_{t\in\mathcal{T}_i}
\log p_{\theta}
\left(
y_{i,t}
\mid q_i,U_i,G_{\pi(i)},y_{i,<t}
\right),
\]

\[
\ell_i^-
=
\frac{s_i^-}{n_i}.
\]

Here, \(\ell_i^+\) corresponds to the matched \(G_i\), while \(\ell_i^-\) corresponds to the mismatched \(G_{\pi(i)}\). The CE-space pairwise DPO objective requires the matched \(G\) tokens to achieve a lower answer CE than the mismatched memory:

\[
\mathcal{L}_{\mathrm{DPO}}
=
\frac{1}{B}
\sum_{i=1}^{B}
\operatorname{softplus}
\left[
-\beta
\left(
\ell_i^- - \ell_i^+
\right)
\right],
\qquad
\beta=0.1.
\]

\paragraph{shuffle-\(G\) KL loss.}

We compute an answer-token-level forward KL only on the shuffle-\(G\) branch. The reference branch disables the generation-memory injection modules and understanding-pathway LoRA and uses the frozen native model conditioned on \(q_i\) and \(U_i\) to generate the reference distribution:

\[
p_{\mathrm{ref},i,t}
=
p_{\mathrm{ref}}
\left(
\cdot\mid q_i,U_i,y_{i,<t}
\right).
\]

The current model's output distribution on the shuffle-\(G\) branch is:

\[
p_{\theta,i,t}^{\mathrm{shuffle}}
=
p_{\theta}
\left(
\cdot
\mid q_i,U_i,G_{\pi(i)},y_{i,<t}
\right).
\]

The shuffle-\(G\) KL loss is therefore defined as:

\[
\mathcal{L}_{\mathrm{KL}}^{\mathrm{shuffle}}
=
\frac{1}{B}
\sum_{i=1}^{B}
\frac{1}{n_i}
\sum_{t\in\mathcal{T}_i}
D_{\mathrm{KL}}
\left(
p_{\mathrm{ref},i,t}
\;\middle\|\;
p_{\theta,i,t}^{\mathrm{shuffle}}
\right).
\]

The reference distribution is detached from the computational graph, and the reference branch is not updated. This term limits the output-distribution shift caused by mismatched \(G\) tokens and prevents the model from enlarging the positive--negative CE gap merely by degrading the shuffle-\(G\) branch.

\paragraph{Gradient decomposition.}

During training, we separately compute the gradients from the real/positive branch and the shuffle-\(G\) branch. The real/positive gradient contains the answer-CE gradient and the positive DPO gradient, while the shuffle-\(G\) branch produces the negative DPO gradient. The positive and negative DPO gradients are combined with weights of \(4:1\), respectively.

The global norm of the negative gradient is then capped at \(0.25\) times that of the real/positive gradient, after which it is added to the real/positive gradient and the shuffle-\(G\) KL gradient. Finally, the combined gradient is clipped to a maximum global norm of \(1.0\).

\paragraph{Understanding-representation masking.}

To reduce reliance on the native understanding pathway during early training, we mask the complete understanding representation \(U_i\) for each sample with probability \(p_U(t)\), while keeping the corresponding generation memory \(G_i\), question, and answer unchanged.
For the formal training run with 5,000 optimizer steps, the mask probability follows the schedule:

\[
p_U(t)=
\begin{cases}
0.8, & 0<t\leq1250,\\
0.5, & 1250<t\leq2500,\\
0.2, & 2500<t\leq3750,\\
0, & 3750<t\leq5000.
\end{cases}
\]

Thus, the model relies more heavily on the \(G\) tokens in the early stages and gradually restores the complete understanding representation, allowing the two pathways to cooperate in the later stages of training.

\section{Experimental Details}
\label{app:experimental-details}

\subsection{Data}

The compression and alignment stage uses 1M images sampled with stratification from Cambrian. For task-conditioned fine-tuning, we select 1,344,016 samples with relevance scores of at least 3 from the non-text-only subset of FineVision. VSR-Bench uses 7,680 training samples from its official random split and 2,195 samples from its official test split. For the medical-domain experiments, we use 19,654 training samples and 6,719 test samples from the official PathVQA splits.

Representative examples from the three datasets are shown below. They illustrate fine-grained multi-object reasoning in FineVision, local spatial-relation recognition in VSR-Bench, and pathological visual question answering in PathVQA.

\begin{figure}[!htbp]
\centering
\setlength{\fboxsep}{7pt}
\setlength{\fboxrule}{0.4pt}

\fbox{%
\begin{minipage}{0.975\linewidth}
\small

\textbf{Training Data Example 1 (FineVision)}\par
\medskip\hrule\medskip

\centering
\begin{minipage}{0.42\linewidth}\centering
  \includegraphics[width=\linewidth]{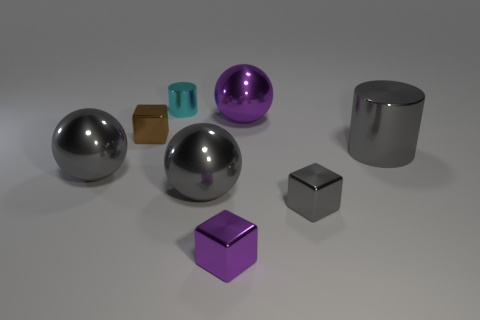}\\[-1mm]
  {\scriptsize \textbf{FineVision: Multi-object reasoning}}
\end{minipage}

\vspace{0.8em}
\hrule
\vspace{0.8em}

{\ttfamily\raggedright\setlength{\parindent}{0pt}\setlength{\parskip}{3pt}%
\begin{tabular}{@{}p{\linewidth}@{}}

\textbf{Question:} Is the number of large things that are in front of the large gray cylinder greater than the number of cyan metal cylinders behind the small brown thing?\\

\textbf{Answer:} Yes.\\

\end{tabular}%
}

\medskip\hrule
\end{minipage}%
}

\caption{Fine-grained visual reasoning example from FineVision.}
\label{fig:appendix_finevision_example}
\end{figure}

\begin{figure}[!htbp]
\centering
\setlength{\fboxsep}{7pt}
\setlength{\fboxrule}{0.4pt}

\fbox{%
\begin{minipage}{0.975\linewidth}
\small

\textbf{Training Example 2 (VSR)}\par
\medskip\hrule\medskip

\centering
\begin{minipage}{0.42\linewidth}\centering
  \includegraphics[width=\linewidth]{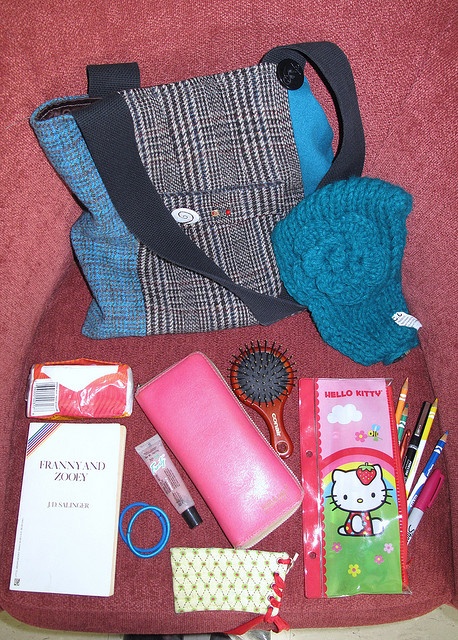}\\[-1mm]
  {\scriptsize \textbf{VSR: Local object relation}}
\end{minipage}

\vspace{0.8em}
\hrule
\vspace{0.8em}

{\ttfamily\raggedright\setlength{\parindent}{0pt}\setlength{\parskip}{3pt}%
\begin{tabular}{@{}p{\linewidth}@{}}

\textbf{Question:} Is the handbag touching the book?\\

A.\ True. \quad B.\ False.\\[2pt]

\textbf{Answer:} B\\

\end{tabular}%
}

\medskip\hrule
\end{minipage}%
}

\caption{Fine-grained local spatial-relation example from VSR.}
\label{fig:appendix_vsr_example}
\end{figure}

\begin{figure}[!htbp]
\centering
\setlength{\fboxsep}{7pt}
\setlength{\fboxrule}{0.4pt}

\fbox{%
\begin{minipage}{0.975\linewidth}
\small

\textbf{Training Example 3 (PathVQA)}\par
\medskip\hrule\medskip

\centering
\begin{minipage}{0.42\linewidth}\centering
  \includegraphics[width=\linewidth]{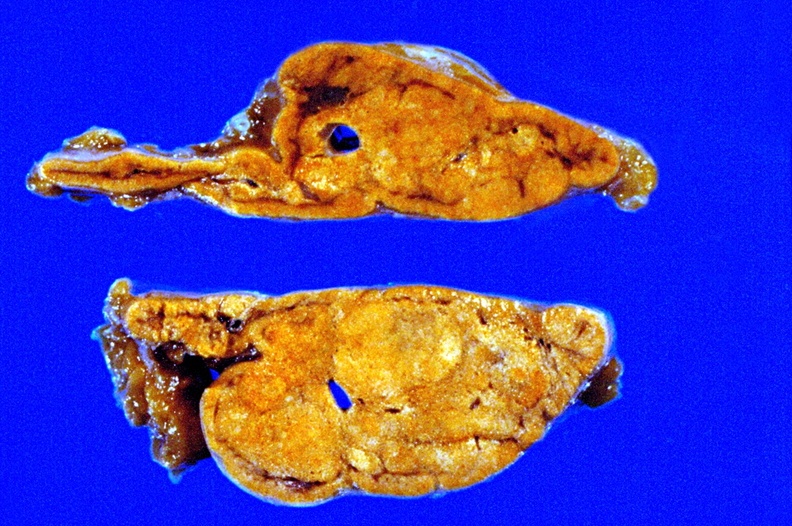}\\[-1mm]
  {\scriptsize \textbf{PathVQA: Pathological detail}}
\end{minipage}

\vspace{0.8em}
\hrule
\vspace{0.8em}

{\ttfamily\raggedright\setlength{\parindent}{0pt}\setlength{\parskip}{3pt}%
\begin{tabular}{@{}p{\linewidth}@{}}

\textbf{Question:} What is shown in this close-up view of the cut surface of the apparently non-functional tissue?\\

\textbf{Answer:} Fixed tissue.\\

\end{tabular}%
}

\medskip\hrule
\end{minipage}%
}

\caption{Fine-grained pathological example from PathVQA.}
\label{fig:appendix_pathvqa_example}
\end{figure}

\subsection{Training}

\paragraph{Stage 1: Compression and alignment.}

We use the AdamW optimizer with a peak learning rate of \(10^{-4}\), warmed up linearly over the first 200 optimizer steps. The learning rate then follows a cosine decay schedule to a minimum of \(10^{-5}\). We set the weight decay to 0.01, \(\beta=(0.9,0.95)\), \(\epsilon=10^{-8}\), and the maximum gradient norm to 1.0. Training uses 8 GPUs, with a micro-batch size of 4 per GPU, an effective global batch size of 32, and 8,000 optimizer steps.

\paragraph{Stage 2: Task-conditioned injection and fine-tuning.}

Training uses an effective global batch size of 32 for 5,000 optimizer steps. The injection projector and internal attention modules use a peak learning rate of \(2\times10^{-5}\), while the understanding-pathway LoRA uses \(10^{-4}\). Both learning rates are linearly warmed up over the first 500 optimizer steps and then follow cosine decay to \(2\times10^{-6}\) and \(10^{-5}\), respectively. The LoRA rank and alpha are both set to 16. We use AdamW with a weight decay of 0.

\begin{figure}[t]
    \centering
    \includegraphics[
        width=\linewidth
    ]{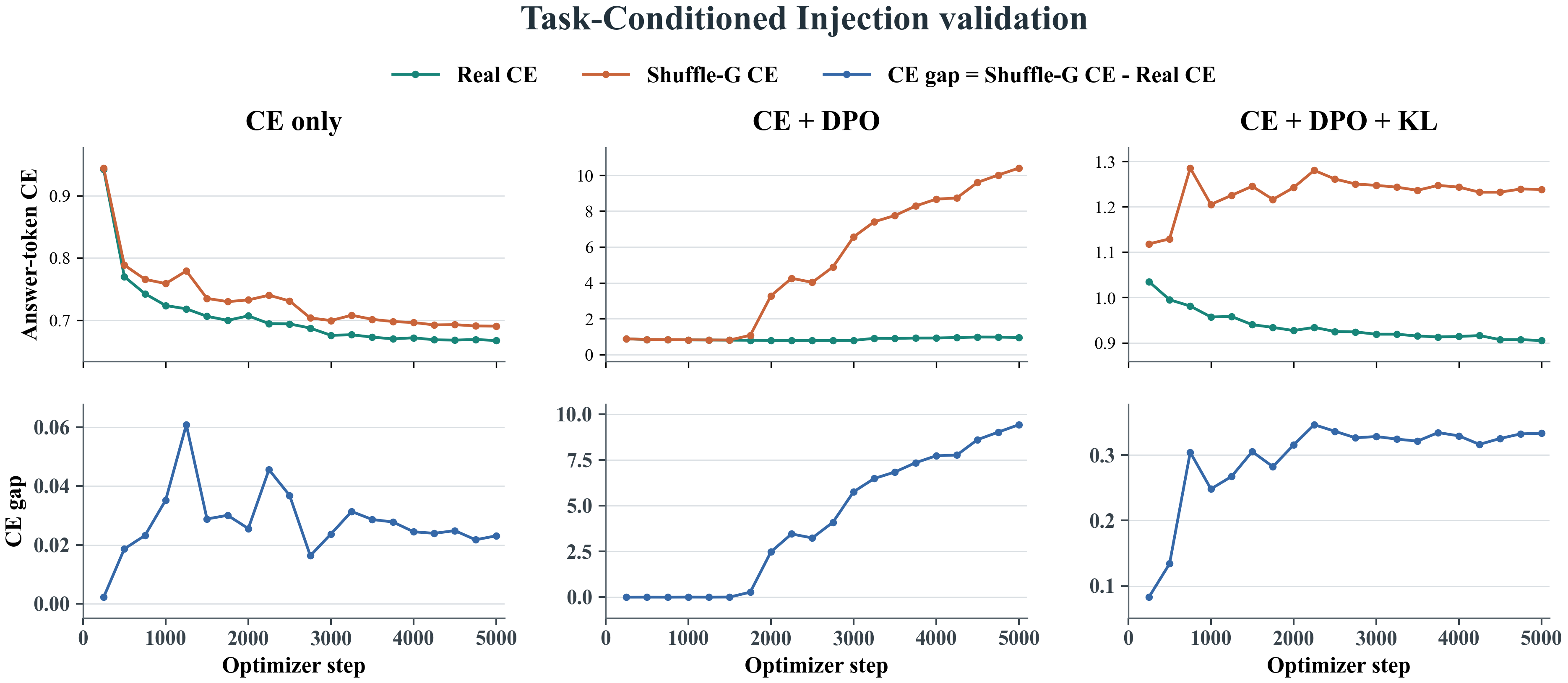}
    \caption{Stage 2 validation dynamics under different fine-tuning objectives.
    The three columns correspond to CE only, CE + DPO, and CE + DPO + KL,
    respectively. The first row reports the answer-token CE of the real and
    shuffle-\(G\) branches, while the second row reports their CE gap,
    defined as \(\mathrm{CE}_{\mathrm{shuffle}-G}
    -\mathrm{CE}_{\mathrm{real}}\). The real branch uses matched \(G\)
    tokens, whereas the shuffle-\(G\) branch uses mismatched \(G\) tokens
    from another sample. A positive CE gap indicates that the model assigns
    lower answer-token cross-entropy to matched \(G\) tokens.}
    \label{fig:stage2-validation-dynamics}
\end{figure}

To verify whether the model uses sample-specific \(G\) tokens, we periodically evaluate the real branch and the shuffle-\(G\) branch on a fixed validation subset. The real branch uses the matched generation memory \(G_i\), whereas the shuffle-\(G\) branch replaces it with \(G_{\pi(i)}\) while keeping the question, answer, and native understanding representation unchanged. Figure~\ref{fig:stage2-validation-dynamics} reports the answer-token CE of the real and shuffle-\(G\) branches and their CE gap under three fine-tuning objectives: CE only, CE + DPO, and CE + DPO + KL. The CE gap is defined as
\(\mathrm{CE}_{\mathrm{shuffle}-G}-\mathrm{CE}_{\mathrm{real}}\).


The CE-only configuration maintains only a small positive CE gap, indicating that the model makes limited use of sample-specific \(G\) tokens. Introducing DPO explicitly optimizes the CE difference between matched and mismatched generation memory, encouraging the model to assign lower answer-token CE to the matched \(G_i\) and thereby increasing the CE gap. However, this objective may enlarge the gap by excessively degrading the shuffle-\(G\) branch. The shuffle-\(G\) KL regularization addresses this issue by constraining the output distribution of the shuffled branch to remain close to that of the frozen reference model, protecting the shuffled branch from uncontrolled deviation while preserving the positive CE gap. Thus, DPO promotes the use of sample-specific generation information, whereas KL regularization controls the cost of using mismatched generation memory.


\subsection{Evaluation}

The evaluation protocols for the benchmarks are as follows:

\begin{itemize}
\item \textbf{VStar-Bench}: We use 191 multiple-choice samples and shuffle the option positions in the formal manifest. For each candidate answer, we compute the average answer-token CE and select the candidate with the lowest CE. Option labels are normalized to uppercase A--D, and we report choice accuracy.

\item \textbf{OmniSpatial}: We use 1,533 spatial-reasoning samples. The main protocol uses forced-choice scoring based on the average candidate-token CE. For each candidate answer, we compute the average token CE and select the candidate with the lowest CE.

\item \textbf{MMVP}: We use 300 questions forming 150 complementary question pairs. For each candidate answer, we compute the average token CE and select the candidate with the lowest CE.

\item \textbf{VSR-Bench}: We use 2,195 samples from the random test split and apply the True/False greedy-generation protocol.

\item \textbf{PathVQA}: We use all 6,719 samples from the official test set for medical visual question answering. We use greedy decoding without sampling until EOS is generated or the maximum number of new tokens is reached. We report normalized exact-match accuracy, where generated and reference answers are lowercased and stripped of punctuation before exact matching.
\end{itemize}

\section{Limitations}
First, GAFU currently targets UMMs with mainstream dual-pathway visual encoders and
does not cover all UMM architectures.
Its application to models with different visual pathway designs or incompatible
representation interfaces requires additional adaptation.
Second, the effectiveness of GAFU depends on the quality of the
native generation representations, and the resulting benefit may vary across models and tasks.
Third, despite its performance gains, extracting, projecting, and injecting the
generation tokens introduces extra inference computation and memory overhead
compared with using the native understanding pathway alone.

\end{document}